\documentclass[11pt]{article}

\usepackage[final]{acl}
\usepackage{times}
\usepackage{latexsym}
\usepackage{tabularx}
\usepackage{array}
\usepackage{kotex}
\usepackage{dblfloatfix}
\usepackage[T1]{fontenc}
\usepackage[utf8]{inputenc}
\usepackage{tcolorbox}
\usepackage{microtype}
\usepackage{inconsolata}
\usepackage{graphicx,hyperref}

\title{Ko-PIQA: A Korean Physical Commonsense Reasoning Dataset \\ with Cultural Context

}

\author{Dasol Choi$^{1,2,5*\dagger}$, Jungwhan Kim$^{3*}$, Guijin Son$^{4,5}$\\
  $^{1}$Yonsei University \quad
  $^{2}$AIM Intelligence \quad
  $^{3}$NAVER Cloud \quad
  $^{4}$OnelineAI \quad
  $^{5}$MODULABS\\
  \texttt{dasolchoi@yonsei.ac.kr, jungwhan.kim@navercorp.com}
}

\begin{document}

\maketitle

\renewcommand{\thefootnote}{\fnsymbol{footnote}}
\footnotetext[1]{\small Equal contribution.}
\footnotetext[2]{\small Corresponding author.}
\setcounter{footnote}{0}
\renewcommand{\thefootnote}{\arabic{footnote}}

\begin{abstract}
Physical commonsense reasoning datasets like PIQA are predominantly English-centric and lack cultural diversity. We introduce Ko-PIQA, a Korean physical commonsense reasoning dataset that incorporates cultural context. Starting from 3.01 million web-crawled questions, we employed a multi-stage filtering approach using three language models to identify 11,553 PIQA-style questions. Through GPT-4o refinement and human validation, we obtained 441 high-quality question-answer pairs.
A key feature of Ko-PIQA is its cultural grounding: 19.7\% of questions contain culturally specific elements like traditional Korean foods (kimchi), clothing (hanbok), and specialized appliances (kimchi refrigerators) that require culturally-aware reasoning beyond direct translation. We evaluate seven language models on Ko-PIQA, with the best model achieving 83.22\% accuracy while the weakest reaches only 59.86\%, demonstrating significant room for improvement. Models particularly struggle with culturally specific scenarios, highlighting the importance of culturally diverse datasets.
Ko-PIQA serves as both a benchmark for Korean language models and a foundation for more inclusive commonsense reasoning research. The dataset and code will be publicly available.
\end{abstract}

\section{Introduction}

Physical commonsense reasoning, the ability to understand how objects interact in the physical world, is fundamental to human intelligence and critical for developing AI systems that can operate effectively in real-world environments \cite{bisk2020piqa, xue2023phy}. While humans effortlessly navigate everyday physical interactions, current AI systems still struggle with seemingly basic physical reasoning tasks that require understanding of material properties, spatial relationships, and cause-and-effect dynamics \cite{sharma2025commonsense}.

The Physical Interaction Question Answering (PIQA) dataset \cite{bisk2020piqa} represents a significant benchmark for evaluating physical commonsense reasoning in English, containing over 20,000 examples that test models' understanding of physical interactions. However, existing physical commonsense datasets like PIQA are predominantly English-centric and fail to capture the cultural diversity that influences how physical reasoning manifests across different societies.

This limitation is particularly pronounced for languages like Korean, where cultural practices, traditional tools, and everyday objects differ significantly from Western contexts. For instance, understanding how to properly care for traditional Korean clothing (hanbok), manage traditional heating systems (ondol), use specialized appliances like kimchi refrigerators, or handle fermented foods requires culturally-aware physical reasoning that cannot be simply translated from English datasets.

We introduce Ko-PIQA,\footnote{Dataset available at 
\href{https://huggingface.co/datasets/HAERAE-HUB/Ko-PIQA}{\texttt{HAERAE-HUB/Ko-PIQA}}.} 
a Korean physical commonsense reasoning dataset that addresses these limitations. 
Through rigorous multi-stage filtering of 3.01 million web-crawled questions and human validation, we created 441 high-quality question-answer pairs, with 87 questions (19.7\%) featuring distinctly Korean cultural elements. Our evaluation of seven language models reveals that current models struggle significantly with Korean physical reasoning, particularly in culturally grounded scenarios, with performance ranging from 59.86\% to 83.22\%. These results highlight the importance of diverse, culturally-aware benchmarks for developing more inclusive AI systems.

\begin{figure*}[t]
    \centering
    \includegraphics[width=0.98\linewidth]{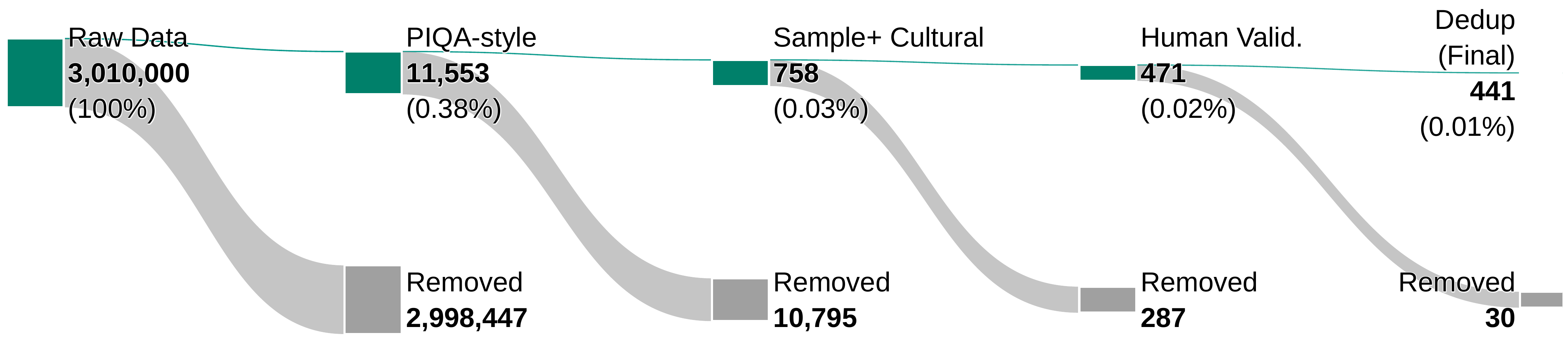}
    \caption{Count-based filtering funnel for Ko-PIQA. Each block shows the remaining number of items after the corresponding filtering step, 
    with percentages relative to the raw set.}
    \label{fig:kopiqa-funnel}
\end{figure*}

\section{Related Work}

The PIQA dataset~\cite{bisk2020piqa} pioneered physical commonsense reasoning, where models must choose between two solutions to daily physical problems. While humans achieve 95\% accuracy, large pretrained models reach only around 77\% performance. Recent advances include Cosmos-Reason1 \cite{azzolini2025cosmos} for multimodal physical reasoning and VideoPhy \cite{videophy2024} for evaluating physical commonsense in video generation. The broader commonsense reasoning field has developed over 139 benchmarks across modalities \cite{commonsense2024survey}, including CommonsenseQA \cite{talmor2018commonsenseqa}, SWAG \cite{zellers2018swag}, and WinoGrande \cite{sakaguchi2021winogrande}.

Korean NLP has established several key benchmarks. KorNLI and KorSTS \cite{ham2020kornli} pioneered Korean natural language understanding through machine translation and manual post-editing of English datasets. KoCommonGEN v2 \cite{seo2024kocommongen} revealed that models struggle with Korean commonsense tasks, with GPT-4 achieving only 74\% accuracy compared to 85\% human performance. KoBBQ \cite{kobbq2024} highlighted how cultural biases depend heavily on cultural context and cannot be simply adapted from English benchmarks.

Recent work emphasizes cultural adaptation in dataset creation, recognizing that direct translation fails to capture culture-specific reasoning patterns \cite{putri2024can, liu2025culturally}. Benchmarks such as EPiK \cite{everydayphysicskoreancontexts} further demonstrate its need in the Korean context. Our work addresses this gap by creating the first Korean physical commonsense reasoning dataset that combines multilingual adaptation with culturally grounded scenarios, filling a critical need in the current landscape of commonsense reasoning benchmarks.


\begin{figure*}[h]
\centering
\includegraphics[width=\textwidth]
{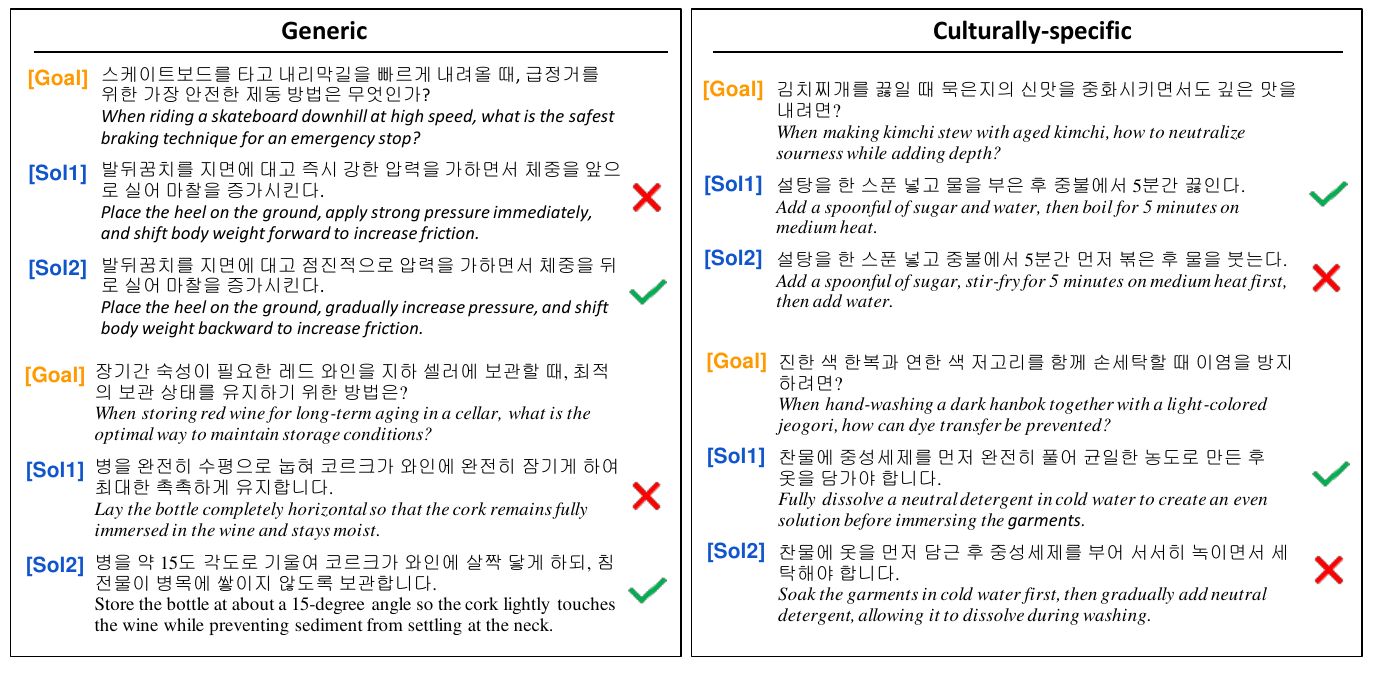}
\caption{Examples from Ko-PIQA showing universal (left) and culturally-specific (right) physical reasoning questions. Check marks indicate correct answers.}
\label{fig:example}
\vspace{-0.8em}
\end{figure*}

\section{Ko-PIQA Dataset}

We construct Ko-PIQA through a multi-stage process that filters, samples, 
and validates questions from 3.01M raw Q\&A pairs to yield 441 high-quality PIQA-style items. 
Figure~\ref{fig:kopiqa-funnel} provides an overview of the entire pipeline,
showing the remaining counts and percentages at each step.
Figure~\ref{fig:example} further illustrates representative Ko-PIQA items,
including both general physical commonsense questions and culturally-grounded examples
that require Korea-specific knowledge.

\smallskip
\noindent\textbf{Data Collection.} We collected 3.01 million Korean questions from Naver Knowledge iN\footnote{\url{https://kin.naver.com}}, a popular Korean Q\&A platform, spanning data up to May 2025. These questions cover diverse everyday scenarios where Korean users seek practical advice on physical tasks and problem-solving.

\smallskip
\noindent\textbf{Multi-Model Filtering.} To identify PIQA-style questions, we employed three language models with diverse architectures and scales: Qwen3-4B \cite{yang2025qwen3}, Qwen3-32B \cite{yang2025qwen3}, and HCX-14B \cite{team2025hyperclova}, a Korean-specialized model. This combination ensures robust filtering by leveraging both multilingual models (Qwen) and a Korean-centric model (HCX) across different parameter scales (4B to 32B). Each model classified questions using a detailed prompt that defined PIQA-style characteristics: requiring physical commonsense about concrete procedures, tool usage, material manipulation, and practical tasks (e.g., repair, cleaning, storage), while excluding pure knowledge questions, abstract concepts, or non-physical advice. We retained only questions that all three models unanimously classified as PIQA-style, yielding 11,553 candidates from the original 3.01 million.

\smallskip
\noindent\textbf{Sampling and Refinement.} From the 11,553 PIQA-style questions, we sampled 600 general physical commonsense questions and separately identified 158 questions containing Korean cultural elements (e.g., references to kimchi, hanbok, ondol). We included both sets, resulting in 758 questions for refinement. Using GPT-4o \cite{hurst2024gpt}, we refined these questions and generated challenging distractors, which are plausible but incorrect alternatives that require careful physical reasoning to eliminate.

\smallskip
\noindent\textbf{Human Validation.} Two native Korean speakers independently validated each question. They filtered out items that were trivial, inappropriate, or required specialized expertise beyond everyday knowledge. The annotators also improved question clarity, calibrated difficulty levels, and verified cultural appropriateness, reducing the dataset to 471 high-quality questions. 

\smallskip
\noindent\textbf{Deduplication.} Using KoSentenceBERT \cite{kosentencebert}, we removed near-duplicate questions with cosine similarity exceeding 0.85. This threshold eliminated redundancy while preserving topically related but distinct questions, yielding our final dataset of 441 unique question-answer pairs.


\subsection{Dataset Analysis}

\textbf{Dataset Statistics.} Ko-PIQA contains 441 question-answer pairs, each consisting of a goal statement and two candidate solutions with exactly one correct answer. Questions average 66.1 characters (median: 64, range: 26-139), while answers average 62.3 characters. The dataset maintains perfect label balance (50.1\% label 1, 49.9\% label 0), ensuring unbiased evaluation. All questions follow a binary choice format consistent with the original PIQA structure, enabling direct comparison of model performance across languages.

\smallskip
\noindent\textbf{Cultural Specificity Analysis.} A distinctive feature of Ko-PIQA is its cultural grounding: 87 questions (19.7\%) contain elements specific to Korean culture that require contextual knowledge beyond direct translation. These include traditional foods and cooking methods (kimchi preparation, makgeolli brewing), traditional clothing care (hanbok washing), housing systems (ondol heating), specialized appliances (kimchi refrigerators), cultural practices (traditional music instruments), and location-specific scenarios (Korean geography and urban environments). The remaining 354 questions (80.3\%) represent universal physical reasoning scenarios adapted to Korean language and context. This balance ensures the dataset serves both as a Korean language benchmark and a test of cultural adaptation in AI systems.


\section{Experimental Setup}

We evaluate seven language models on Ko-PIQA to assess their Korean physical commonsense reasoning capabilities. Our evaluation includes models of varying sizes and architectures: Qwen3-32B and Qwen3-8B \cite{yang2025qwen3}, Gemma3-27B-it and Gemma3-9B-it \cite{team2025gemma}, KANANA-1.5-8B-instruct-2505 \cite{bak2025kanana}, Meta-Llama-3-8B-Instruct \cite{dubey2024llama}, and EXAONE-4.0-7.8B-Instruct \cite{research2025exaone}. Notably, KANANA and EXAONE are Korean-specialized models, while others are multilingual models with Korean capabilities.
All models are evaluated using zero-shot prompting in Korean with the following template:

\begin{tcolorbox}
[colback=gray!5!white,colframe=gray!75!black,boxrule=0.5pt]
\small
다음 질문에 대해 두 개의 답변 중 어느 것이 더 정확하고 도움이 되는지 판단해주세요. 
정답은 반드시 A 또는 B 중 하나로만 출력해주세요.\\
\textit{(Please determine which of the two answers is more accurate and helpful 
for the following question. You must answer with either A or B only.)}\\[0.5em]
\textbf{출력 형식 예시 (\textit{Output Format Example}):} \\
정답 (Answer): A\\[0.5em]
\textbf{질문 (\textit{Question})}: [goal]\\
\textbf{답변 A (\textit{Answer A})}: [solution 0]\\
\textbf{답변 B (\textit{Answer B})}: [solution 1]
\small
\end{tcolorbox}

Evaluation is conducted using exact match accuracy, where a response is considered correct only if it exactly matches the ground truth label.
We implemented the evaluation using the \texttt{lm-eval-harness} framework \cite{eval_harness} with the task set as \texttt{generation}, and post-processed model outputs to compute binary exact-match accuracy (equivalent to classification accuracy).
This ensures deterministic scoring and reproducibility across models. 
We report overall accuracy across all 441 questions and separately analyze performance on cultural-specific questions (87 questions) versus universal questions (354 questions). All evaluations were conducted in non-thinking mode.

\section{Results}

Table~\ref{tab:results} presents the overall performance of seven language models on Ko-PIQA. The results reveal significant challenges in Korean physical commonsense reasoning, with performance ranging from 59.86\% to 83.22\%.

\begin{table}[h]
\centering
\renewcommand{\arraystretch}{0.9} 
\begin{tabular}{lcc}
\hline
\textbf{Model} & \textbf{Overall} & \textbf{Cultural} \\
\hline
\multicolumn{3}{l}{\textit{Multilingual Models}} \\
Qwen3-32B & 83.22 & 86.21 \\
Qwen3-8B & 71.88 & 74.71 \\
Gemma3-27B-it & 79.59 & 79.31 \\
Gemma3-4B-it & 65.53 & 70.11 \\
Meta-Llama-3-8B & 59.86 & 67.82 \\
\hline
\multicolumn{3}{l}{\textit{Korean-specialized Models}} \\
EXAONE-4.0-32B & 81.86 & 87.36 \\
KANANA-1.5-8B & 66.89 & 79.31 \\
\hline
\end{tabular}
\caption{Accuracy (\%) on Ko-PIQA dataset. Cultural refers to the 87 culturally-specific questions.}
\label{tab:results}
\end{table}

\smallskip
\noindent\textbf{Overall Performance.} The best-performing model, Qwen3-32B, achieves 83.22\% accuracy, while the weakest, Meta-Llama-3-8B, reaches only 59.86\%. This substantial performance gap demonstrates that Korean physical commonsense reasoning remains challenging for current language models, with even the strongest models falling short of human-level performance (estimated at 95\% for English PIQA).

\smallskip
\noindent\textbf{Cultural vs. Universal Questions.} Interestingly, most models perform comparably or even better on culturally-specific questions than on universal ones. EXAONE-4.0-32B shows the strongest cultural performance (87.36\%), likely benefiting from its Korean-centric training. However, models like Gemma3-27B-it struggle more with cultural questions (79.31\% vs. 79.59\% overall), suggesting that cultural adaptation requires more than just language proficiency.

\smallskip
\noindent\textbf{Model Size Effects.} Larger models generally outperform smaller ones within the same family (Qwen3-32B > Qwen3-8B, Gemma3-27B > Gemma3-4B), indicating that scale helps with complex reasoning tasks. However, KANANA-1.5-8B's strong cultural performance (79.31\%) despite its smaller size highlights the importance of Korean-specific training data.

\section{Limitations}

While Ko-PIQA addresses the lack of Korean physical commonsense reasoning benchmarks, it has several limitations.  
First, with 441 questions, it is much smaller than PIQA (20K+ examples),
which may limit its utility for model fine-tuning or few-shot training.  
Second, although we include 87 culturally-specific items (19.7\%),
this does not cover the full breadth of Korean cultural practices.  
Finally, we only report zero-shot model performance and leave few-shot or fine-tuning
evaluation for future work.

\section{Conclusion}

We present Ko-PIQA, the first Korean physical commonsense reasoning dataset with 441 question-answer pairs, including 87 culturally-specific questions (19.7\%). Our evaluation reveals that current language models achieve 59.86-83.22\% accuracy, with Korean-centric models showing superior cultural understanding. Ko-PIQA demonstrates that effective physical reasoning requires cultural knowledge beyond simple translation, providing a valuable benchmark for developing culturally-aware AI systems. The dataset will be publicly available.

\section*{Acknowledgments}
This research was supported by Brian Impact, a non-profit organization dedicated to advancing science and technology.

\bibliography{custom}

\end{document}